\documentclass[final]{cvpr}

\usepackage{times}
\usepackage{epsfig}
\usepackage{graphicx}
\usepackage{amsmath}
\usepackage{amssymb}
\usepackage{booktabs}
\usepackage{multirow}
\usepackage{color}
\usepackage{tabularx}
\usepackage{cite}

\usepackage[pagebackref=true,breaklinks=true,colorlinks,bookmarks=false]{hyperref}

\def\cvprPaperID{7804} 
\def\confYear{CVPR 2021}

\newcommand{\tabincell}[2]{\begin{tabular}{@{}#1@{}}#2\end{tabular}}
\begin{document}

\title{Generating Diverse Structure for Image Inpainting With Hierarchical VQ-VAE}

\author{Jialun Peng$^1\qquad\!$ Dong Liu$^1$\thanks{This work was supported by the Natural Science Foundation of China under Grants 62036005 and 62022075. \emph{(Corresponding author: Dong Liu.)}}$\qquad\!$ Songcen Xu$^2\qquad\!$ Houqiang Li$^1$\\
$^1$ University of Science and Technology of China $\ \ ^2$ Noah’s Ark Lab, Huawei Technologies Co., Ltd.\\
{\tt\small pjl@mail.ustc.edu.cn, \{dongeliu, lihq\}@ustc.edu.cn, xusongcen@huawei.com}
}

\maketitle


\begin{abstract}

	Given an incomplete image without additional constraint, image inpainting natively allows for multiple solutions as long as they appear plausible. Recently, multiple-solution inpainting methods have been proposed and shown the potential of generating diverse results. However, these methods have difficulty in ensuring the quality of each solution, e.g.\ they produce distorted structure and/or blurry texture. We propose a two-stage model for diverse inpainting, where the first stage generates multiple coarse results each of which has a different structure, and the second stage refines each coarse result separately by augmenting texture. The proposed model is inspired by the hierarchical vector quantized variational auto-encoder (VQ-VAE), whose hierarchical architecture disentangles structural and textural information. In addition, the vector quantization in VQ-VAE enables autoregressive modeling of the discrete distribution over the structural information. Sampling from the distribution can easily generate diverse and high-quality structures, making up the first stage of our model. In the second stage, we propose a structural attention module inside the texture generation network, where the module utilizes the structural information to capture distant correlations. We further reuse the VQ-VAE to calculate two feature losses, which help improve structure coherence and texture realism, respectively. Experimental results on CelebA-HQ, Places2, and ImageNet datasets show that our method not only enhances the diversity of the inpainting solutions but also improves the visual quality of the generated multiple images. Code and models are available at: \url{https://github.com/USTC-JialunPeng/Diverse-Structure-Inpainting}.
\end{abstract}

\vspace{-3mm}
\section{Introduction}
\begin{figure}[t]
	\begin{center}
		\includegraphics[width=1.0\linewidth]{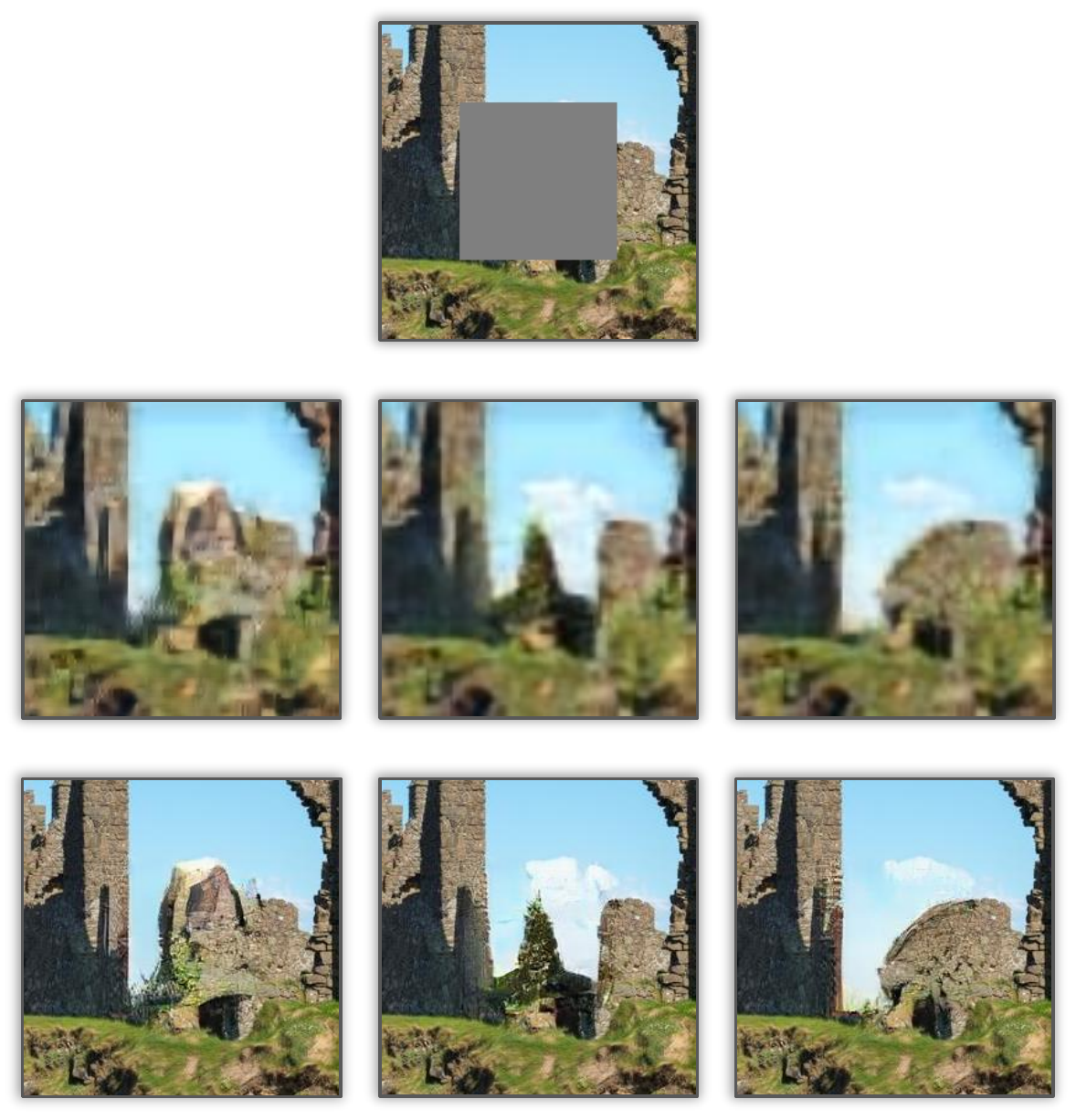}
	\end{center}
	\vspace{-0.3cm}
	\caption{(Top) Input incomplete image, where the missing region is depicted in gray. (Middle) Visualization of the generated diverse structures. (Bottom) Output images of our method.}
	\label{fig1}
	\vspace{-0.3cm}
\end{figure}

Image inpainting refers to the task of filling in the missing region of an incomplete image so as to produce a complete and visually plausible image. Inpainting benefits a series of applications including object removal, photo restoration, and transmission error concealment. As an ill-posed problem, inpainting raises a great challenge especially when the missing region is large and contains complex content. As such, inpainting has attracted much research attention.


Recently, a series of deep learning-based methods are proposed for inpainting \cite{pathak2016context, iizuka2017globally}. They usually employ encoder-decoder architectures and train the networks with the combinations of reconstruction and adversarial losses. To enhance the visual quality of the results, a number of studies \cite{song2018contextual, yan2018shift, yu2018generative, sagong2019pepsi, zeng2019learning} adopt the contextual attention mechanisms to use the available content for generating the missing content. Also, several studies \cite{liu2018image, yu2019free, yi2020contextual} propose modified convolutions in replacement of normal convolutions to reduce the artifacts.


The aforementioned methods all learn a deterministic mapping from an incomplete image to a complete image. However, in practice, the solution to inpainting is not unique. Without additional constraint, multiple inpainting results are equally/similarly plausible for an incomplete image, especially when the missing region is large and contains complex content (\eg Figure \ref{fig1}). Moreover, for typical applications, providing multiple inpainting results may enable the user to select from them according to his/her own preference. It then motivates the design for multiple-solution inpainting.

In contrast to single-solution methods, multiple-solution inpainting shall build a probabilistic model of the missing content conditioned on the available content. Several recent studies \cite{zheng2019pluralistic, zhao2020uctgan} employ variational auto-encoder (VAE) architectures and train the networks with the combinations of Kullback-Leibler (KL) divergences and adversarial losses. VAE-based methods \cite{zheng2019pluralistic, zhao2020uctgan} assume a Gaussian distribution over continuous latent variables. Sampling from the Gaussian distribution presents diverse latent features and leads to diverse inpainted images. Although these methods can generate multiple solutions, some of their solutions are of low quality due to distorted structures and/or blurry textures. It may be attributed to the limitation of the parametric (\eg Gaussian) distribution when we try to model the complex natural image content. In addition, recent studies more and more demonstrate the importance of structural information, \eg segmentation maps \cite{song2018spg, liao2020guidance}, edges \cite{nazeri2019edgeconnect, xiong2019foreground}, and smooth images \cite{ren2019structureflow, liu2020rethink}, for guiding image inpainting. Such structural information is yet to be incorporated into multiple-solution inpainting. Thus, VAE-based methods \cite{zheng2019pluralistic, zhao2020uctgan} tend to produce multiple results with limited structural diversity, which is called posterior collapse in \cite{van2017neural}.


In this paper we try to address the limitations of the existing multiple-solution inpainting methods. First, instead of parametric distribution modeling of continuous variables, we resort to autoregressive modeling of discrete variables. Second, we want to generate multiple structures in an explicit fashion, and then base the inpainting upon the generated structure. We find that the hierarchical vector quantized VAE (VQ-VAE) \cite{razavi2019generating} is suitable for our study\footnote{In this paper we use the basic model in \cite{razavi2019generating}, which is called VQ-VAE-2. Note that our method can use other hierarchical VQ-VAE models as well.}. First, there is a vector quantization step in VQ-VAE making the latent variables to be all discrete; as noted in \cite{van2017neural}, these discrete latent variables allow the usage of powerful decoders to avoid the posterior collapse. Second, the hierarchical layout encourages the split of the image information into global and local parts; with proper design, it may disentangle structural features from textural features of an image.

Based on the hierarchical VQ-VAE, we propose a two-stage model for multiple-solution inpainting. The first stage is known as diverse structure generator, where sampling from a conditional autoregressive distribution produces multiple sets of structural features. The second stage is known as texture generator, where an encoder-decoder architecture is used to produce a complete image based on the guidance of a set of structural features. Note that each set of the generated structural features leads to a complete image (see Figure \ref{fig1}).

The main contributions we have made in this paper can be summarized as follows:\vspace{-0.36cm}

\begin{itemize}
	\item[$\bullet$] We propose a multiple-solution image inpainting method based on hierarchical VQ-VAE. The method has two distinctions from previous multiple-solution methods: first, the model learns an autoregressive distribution over discrete latent variables; second, the model splits structural and textural features.\vspace{-0.30cm}
	\item[$\bullet$] We propose to learn a conditional autoregressive network for the distribution over structural features. The network manages to generate reasonable structures with high diversity.\vspace{-0.30cm}
	\item[$\bullet$] For texture generation  we propose a structural attention module to capture distant correlations of structural features. We also propose two new feature losses to improve structure coherence and texture realism.\vspace{-0.30cm}
	\item[$\bullet$] Extensive experiments on three benchmark datasets including CelebA-HQ, Places2, and ImageNet demonstrate the superiority of our proposed method in both quality and diversity.\vspace{-0.10cm}

\end{itemize}


\section{Related Work}
\vspace{-0.05cm}

\subsection{Image Inpainting}
Traditional image inpainting methods such as diffusion-based methods \cite{bertalmio2000image, ballester2001filling} and patch-based methods \cite{bertalmio2003simultaneous, drori2003fragment, criminisi2004region, barnes2009patchmatch} borrow image-level patches from source images to fill in the missing regions. They are unable to generate unique content not found in the source images. Furthermore, these methods often generate unreasonable results without considering high-level semantics of the images.

Recently, learning-based methods which use deep convolutional networks are proposed to semantically predict the missing regions. Pathak \etal \cite{pathak2016context} first apply adversarial learning to the image inpainting. Iizuka \etal \cite{iizuka2017globally} introduce an extra discriminator to enforce the local consistency. Yan \etal \cite{yan2018shift} and Yu \etal \cite{yu2018generative} propose patch-swap and contextual attention to make use of distant feature patches for the higher inpainting quality.  Liu \etal \cite{liu2018image} and Yu \etal \cite{yu2019free} introduce partial convolutions and gated convolutions to reduce visual artifacts caused by normal convolutions. In order to generate reasonable structures and realistic textures, Nazeri \etal \cite{nazeri2019edgeconnect} and Xu \etal \cite{xu2020e2i} use edge maps as structural information to guide image inpainting. Ren \etal \cite{ren2019structureflow} propose to use edge-preserved smooth images instead of edge maps. Liu \etal \cite{liu2020rethink} propose feature equalizations to improve the consistency between the structure and the texture. However, these learning-based methods only generate one optimal result for each incomplete input. They focus on reconstructing ground truth rather than creating plausible results.

To obtain multiple inpainting solutions, Zheng \etal \cite{zheng2019pluralistic} propose a VAE-based model with two parallel paths, which trades off between reconstructing ground truth and maintaining the diversity of the inpainting results. Zhao \etal \cite{zhao2020uctgan} propose a similar VAE-based model which uses instance images to improve the diversity. However, these methods do not effectively separate the structural and textural information, they often produce distorted structures and/or blurry textures.



\subsection{VQ-VAE and Autoregressive Networks}
The vector quantized variational auto-encoder (VQ-VAE) \cite{van2017neural} is a discrete latent VAE model which relies on vector quantization layers to model discrete latent variables. The discrete latent variables allow a powerful autoregressive network such as PixelCNN \cite{oord2016pixel, van2016conditional, salimans2017pixelcnn++, chen2018pixelsnail} to model latents without worrying about the posterior collapse problem \cite{van2017neural}. Razavi \etal \cite{razavi2019generating} propose a hierarchical VQ-VAE which uses a hierarchy of discrete latent variables to separate the structural and textural information. Then they use two PixelCNNs to model structural and textural information, respectively. However, the PixelCNNs are conditioned on the class label for image generation, while there is no class label in the image inpainting task. Besides, the generated textures of the PixelCNNs lack fine-grained details due to the lossy nature of VQ-VAE. It relieves the generation model from modeling negligible information, but it hinders the inpainting model from generating realistic textures consistent with the known regions.
The PixelCNNs are thus not practical for image inpainting.

\section{Method}

\begin{figure*}[t]
	\begin{center}
		\includegraphics[width=1.0\linewidth]{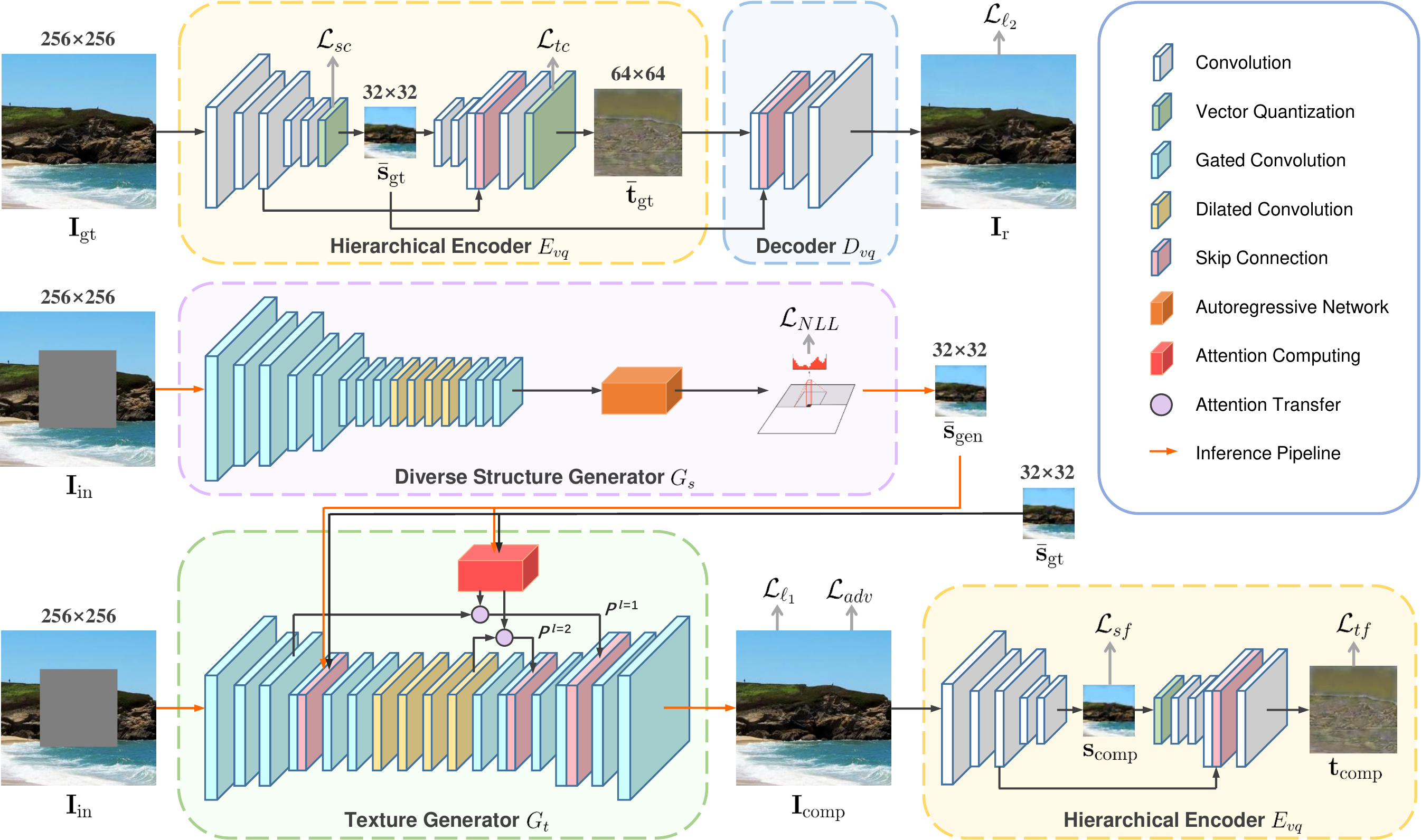}
	\end{center}
	\vspace{-0.2cm}
	\caption{Overview of the proposed method. (Top) Hierarchical vector quantized variational auto-encoder (VQ-VAE) consists of hierarchical encoder $E_{vq}$ and decoder $D_{vq}$. $E_{vq}$ extracts discrete structural features $\bar{\textbf{s}}_{\rm gt}$ and discrete textural features $\bar{\textbf{t}}_{\rm gt}$. $D_{vq}$ reconstructs the image from these two sets of discrete features. (Middle) Diverse structure generator $G_s$ models the conditional distribution over the discrete structural features by using an autoregressive network, where $\bar{\textbf{s}}_{\rm gt}$ is used to calculate the loss $\mathcal{L}_{N\!L\!\!\:L}$. During inference, sampling from the distribution can generate multiple possible structural features $\bar{\textbf{s}}_{\rm gen}$. (Bottom) Texture generator $G_t$ synthesizes the image texture given the discrete structural features ($\bar{\textbf{s}}_{\rm gt}$ in the training and $\bar{\textbf{s}}_{\rm gen}$ in the inference). The pre-trained $E_{vq}$ is used as an auxiliary evaluator to improve image quality, where $\bar{\textbf{s}}_{\rm gt}$ and $\bar{\textbf{t}}_{\rm gt}$ are used to calculate the losses $\mathcal{L}_{sf}$ and $\mathcal{L}_{tf}$. During training, the hierarchical VQ-VAE is firstly trained, and then $G_s$ and $G_t$ are trained individually. During inference, only $G_s$ and $G_t$ are used.}
	\label{fig2}
	\vspace{-0.2cm}
\end{figure*}

As shown in Figure \ref{fig2}, the pipeline of our method consists of three parts: hierarchical VQ-VAE $E_{vq}$-$D_{vq}$, diverse structure generator $G_s$ and texture generator $G_t$. The hierarchical encoder $E_{vq}$ disentangles discrete structural features and discrete textural features of the ground truth ${\rm \mathbf{I}_{gt}}$ and the decoder $D_{vq}$ outputs the reconstructed image ${\rm \mathbf{I}_{r}}$. The diverse structure generator $G_s$ produces diverse discrete structural features given an input incomplete image ${\rm \mathbf{I}_{in}}$. The texture generator $G_t$ synthesizes the image texture given the discrete structural features and outputs the completion result  ${\rm \mathbf{I}_{comp}}$. We also use the pre-trained $E_{vq}$ as an auxiliary evaluator to define two novel feature losses for better visual quality of the completion result.

\subsection{Hierarchical VQ-VAE}
In order to disentangle structural and textural information, we pre-train a hierarchical VQ-VAE $E_{vq}$-$D_{vq}$ following \cite{razavi2019generating}.
The hierarchical encoder $E_{vq}$ maps ground truth ${\rm \mathbf{I}_{gt}}$ onto structural features $\textbf{s}_{\rm gt}$ and textural features $\textbf{t}_{\rm gt}$. The processing of $E_{vq}$ can be written as $(\textbf{s}_{\rm gt}, \textbf{t}_{\rm gt})=E_{vq}({\rm \mathbf{I}_{gt}})$. These features are then quantized to discrete features by two vector quantization layers. Each vector quantization layer has $K$ = 512 prototype vectors in its codebook and the vector dimensionality is $D$ = 64. As such, each vector of features is replaced by the nearest prototype vector based on Euclidean distance. The processing of vector quantization can be written as $\bar{\textbf{s}}_{\rm gt}=VQ_s(\textbf{s}_{\rm gt})$ and $\bar{\textbf{t}}_{\rm gt}=VQ_t(\textbf{t}_{\rm gt})$. Finally, the decoder $D_{vq}$ reconstructs image from these two sets of discrete features. The processing of $D_{vq}$ can be written as ${\rm \mathbf{I}_{r}}=D_{vq}(\bar{\textbf{s}}_{\rm gt},\bar{\textbf{t}}_{\rm gt})$.

The reconstruction loss of $E_{vq}$-$D_{vq}$ is defined as:
\begin{equation}
	\mathcal{L}_{\ell_2}={\|{\rm \mathbf{I}_{r}}-{\rm \mathbf{I}_{gt}}\|}_2^2
\end{equation}
To back-propagate the gradient of the reconstruction loss through vector quantization, we use the straight-through gradient estimator \cite{bengio2013estimating}. The codebook prototype vectors are updated using the exponential moving average of the encoder output. As proposed in \cite{van2017neural}, we also use two commitment losses $\mathcal{L}_{ sc}$ and $\mathcal{L}_{tc}$ to align the encoder output with the codebook prototype vectors for stable training. The commitment loss of structural features is defined as:
\begin{equation}
	\mathcal{L}_{sc}={\|\textbf{s}_{\rm gt}-sg[\,\bar{\textbf{s}}_{\rm gt}]\|}_2^2
\end{equation}
where $sg$ denotes the stop-gradient operator \cite{van2017neural}. The commitment loss of textural features (denoted as $\mathcal{L}_{tc}$) is similar to $\mathcal{L}_{sc}$. The total loss of $E_{vq}$-$D_{vq}$ is defined as:
\begin{equation}
	\mathcal{L}_{vq}=\alpha_{\ell_2}\mathcal{L}_{\ell_2}+\alpha_c(\mathcal{L}_{sc}+\mathcal{L}_{tc})
\end{equation}
where $\alpha_{\ell_2}$ and $\alpha_c$ are loss weights.

For 256$\times$256 images, the size of structural features is 32$\times$32 and that of textural features is 64$\times$64. We visualize their discrete representations using the decoder $D_{vq}$. The visualized results can be written as $D_{vq}(\bar{\textbf{s}}_{\rm gt},\textbf{0})$ and $D_{vq}(\textbf{0},\bar{\textbf{t}}_{\rm gt})$, where $\textbf{0}$ is a zero tensor. As shown in Figure \ref{fig2}, the structural features model global information such as shapes and colors, and the textural features model local information such as details and textures.

\subsection{Diverse Structure Generator}


Previous multiple-solution inpainting methods \cite{zhao2020uctgan,zheng2019pluralistic} often produce distorted structures and/or blurry textures, suggesting that these methods struggle to recover the structure and the texture simultaneously. Therefore, we first propose a diverse structure generator $G_s$ which uses an autoregressive network to formulate a conditional distribution over the discrete structural features. Sampling from the distribution can produce diverse structural features.

Similar to the PixelCNN in \cite{razavi2019generating}, our autoregressive network consists of 20 residual gated convolution layers and 4 casual multi-headed attention layers. Since the PixelCNN in \cite{razavi2019generating} is conditioned on the class label for the image generation task, we make two modifications to make it practical for the image inpainting task. First, we stack gated convolution layers to map the input incomplete image ${\rm \mathbf{I}_{in}}$ and its binary mask ${\rm \mathbf{M}}$ to a condition. The condition is injected into each residual gated convolution layer of the autoregressive network. Second, we use a light-weight autoregressive network by reducing both the hidden units and the residual units to 128 for efficiency.

During training, $G_s$ utilizes the input incomplete image as the condition and models the conditional distribution over $\bar{\textbf{s}}_{\rm gt}$. This distribution can be written as $p_{\theta}(\bar{\textbf{s}}_{\rm gt}|{\rm \mathbf{I}_{in}}, {\rm \mathbf{M}})$, where $\theta$ denotes network parameters of $G_s$.  The training loss of $G_s$ is defined as the negative log likelihood of $\bar{\textbf{s}}_{\rm gt}$:
\begin{equation}
	\mathcal{L}_{N\!L\!\!\:L}=-\,\mathbb{E}_{\,{\rm \mathbf{I}_{gt}}\sim p_{data}}[{\rm log}\,p_{\theta}(\bar{\textbf{s}}_{\rm gt}|{\rm \mathbf{I}_{in}, {\rm \mathbf{M}}})]
\end{equation}
where $p_{data}$ denotes the distribution of training dataset. During inference, $G_s$ utilizes the input incomplete image as condition and outputs a conditional distribution for generating structural features. This distribution can be written as $p_{\theta}(\bar{\textbf{s}}_{\rm gen}|{\rm \mathbf{I}_{in}}, {\rm \mathbf{M}})$. Sampling from $p_{\theta}(\bar{\textbf{s}}_{\rm gen}|{\rm \mathbf{I}_{in}}, {\rm \mathbf{M}})$ sequentially can generate diverse discrete structural features $\bar{\textbf{s}}_{\rm gen}$.

Due to the low-resolution of structural features, our diverse structure generator can better capture global information. It thus helps generate reasonable global structures. In addition, the training objective of our diverse structure generator is to maximize likelihood of all samples in the training set without any additional loss. Thus, the generated structures do not suffer from GAN’s known shortcomings such as mode collapse and lack of diversity.


\subsection{Texture Generator}
\textbf{Network Architecture.} After obtaining the generated structural features $\bar{\textbf{s}}_{\rm gen}$, our texture generator $G_t$ synthesizes the image texture based on the guidance of $\bar{\textbf{s}}_{\rm gen}$. The network architecture of $G_t$ is similar to the refine network in \cite{yi2020contextual}. As shown in Figure \ref{fig2}, the network architecture consists of gated convolutions and dilated convolutions. Unlike existing inpainting methods, our texture generator $G_t$ utilizes the given structural features as guidance. The structural features are not only input to the first few layers of $G_t$, but also input to our structural attention module. The proposed structural attention module borrows distant information based on the correlations of the structural features. It thus ensures that the synthesized texture is consistent with the generated structure.

Let ${\rm \mathbf{I}_{out}}$ denote the output of $G_t$. The final completion result ${\rm \mathbf{I}_{comp}}$ is the output ${\rm \mathbf{I}_{out}}$ with the non-masked pixels directly set to ground truth. During training, $G_t$ takes the ground truth structural features $\bar{\textbf{s}}_{\rm gt}$ as input so that ${\rm \mathbf{I}_{comp}}$ is the reconstruction of ground truth. During inference, $G_t$ takes the generated structural features $\bar{\textbf{s}}_{\rm gen}$ as input so that ${\rm \mathbf{I}_{comp}}$ is the inpainting result.

\textbf{Structural Attention Module.} Attention modules are widely used in the existing image inpainting methods. They generally calculate the attention scores on a low-resolution intermediate feature map of the network. However, due to the lack of direct supervision on the attention scores, the  learned attention is insufficiently reliable \cite{zhou2020learning}. These attention modules may refer to unsuitable features, resulting in poor inpainting quality. To address this problem, we propose a structural attention module which directly calculates the attention scores on the structural features. Intuitively, regions with similar structures should have similar textures. Calculating the attention scores on the structural features can model accurate long-range correlations of structural information, thereby improving the consistency between the synthesized texture and the generated structure.

In addition, the attention modules in the existing image inpainting methods distinguish the foreground features and the background features using the down-sampled mask. This hand-crafted design may incorrectly divide the feature map and produce artifacts in the inpainting result. Therefore, our structural attention module calculates full attention scores on the structural features. Unlike the foreground-background cross attention that only models correlations between the foreground and the background, our full attention learns full correlations regardless of the feature division. It thus maintains the global consistency of the inpainting result. Moreover, our full attention does not increase the amount of calculation compared to the cross attention.

Like \cite{yi2020contextual}, our structural attention module consists of an attention computing step and an attention transfer step. The attention computing step extracts 3$\times$3 patches from the input structural features. Then, the truncated distance similarity score \cite{sagong2019pepsi} between the patches at $(x,y)$ and $(x^{\prime}, y^{\prime})$ is defined as:
\begin{equation}
	\tilde{d}_{(x, y),(x^{\prime}, y^{\prime})}=\tanh (-(\frac{d_{(x, y),(x^{\prime}, y^{\prime})}-m}{\sigma}))
\end{equation}
where $d_{(x, y),(x^{\prime}, y^{\prime})}$ is the Euclidean distance, $m$ and $\sigma$ are the mean value and the standard deviation of $d_{(x, y),(x^{\prime}, y^{\prime})}$. The truncated distance similarity scores are applied by a scaled sotfmax layer to output full attention scores:
\begin{equation}
	s^*_{(x, y),(x^{\prime}, y^{\prime})}={\rm softmax}(\lambda_1\tilde{d}_{(x, y),(x^{\prime}, y^{\prime})})
\end{equation}
where $\lambda_1$ = 50. After obtaining the full attention scores from the structural features, the attention transfer step reconstructs lower-level feature maps ($P^{l}$) by using the full attention scores as weights:
\begin{equation}
	q_{(x^{\prime}, y^{\prime})}^{l}=\sum_{x, y} s^*_{(x, y),(x^{\prime}, y^{\prime})} p_{(x, y)}^{l}
	\vspace{-0.1cm}
\end{equation}
where $l\in (1,2)$ is the layer number and $p_{(x, y)}^{l}$ \vspace{-0.09cm}is the patch of $P^l$, and $q_{(x^{\prime}, y^{\prime})}^{l}$ is the patch of the reconstructed feature map. The size of patches varies according to the size of feature map like \cite{yi2020contextual}. Finally, the reconstructed feature maps supplement the decoder of $G_t$ via skip connections.

\textbf{Training Losses.} The total loss of $G_t$ consists of a reconstruction loss, an adversarial loss, and two feature losses. The reconstruction loss of $G_t$ is defined as:
\begin{equation}
	\mathcal{L}_{\ell_1}={\|{\rm \mathbf{I}_{out}}-{\rm \mathbf{I}_{gt}}\|}_1
\end{equation}
We use SN-PatchGAN \cite{yu2019free} as our discriminator $D_t$. The hinge version of the adversarial loss for $D_t$ is defined as:
\begin{equation}
	\begin{aligned}
		\mathcal{L}_{d}=&\;\mathbb{E}_{\,{\rm \mathbf{I}_{gt}}\sim p_{data}}[{\rm ReLU}(1-D_t({\rm \mathbf{I}_{gt}}))]\\
		&+\mathbb{E}_{\,{\rm \mathbf{I}_{comp}} \sim p_{z}}[{\rm ReLU}(1+D_t({\rm \mathbf{I}_{comp}}))]
	\end{aligned}
\end{equation}
where $p_{z}$ denotes the distribution of the inpainting results. The adversarial loss for $G_t$ is defined as:
\begin{equation}
	\mathcal{L}_{adv}=-\,\mathbb{E}_{\,{\rm \mathbf{I}_{comp}} \sim p_{z}}[D_t({\rm \mathbf{I}_{comp}})]
\end{equation}
\begin{figure*}
	\begin{center}
		\includegraphics[width=0.83\linewidth]{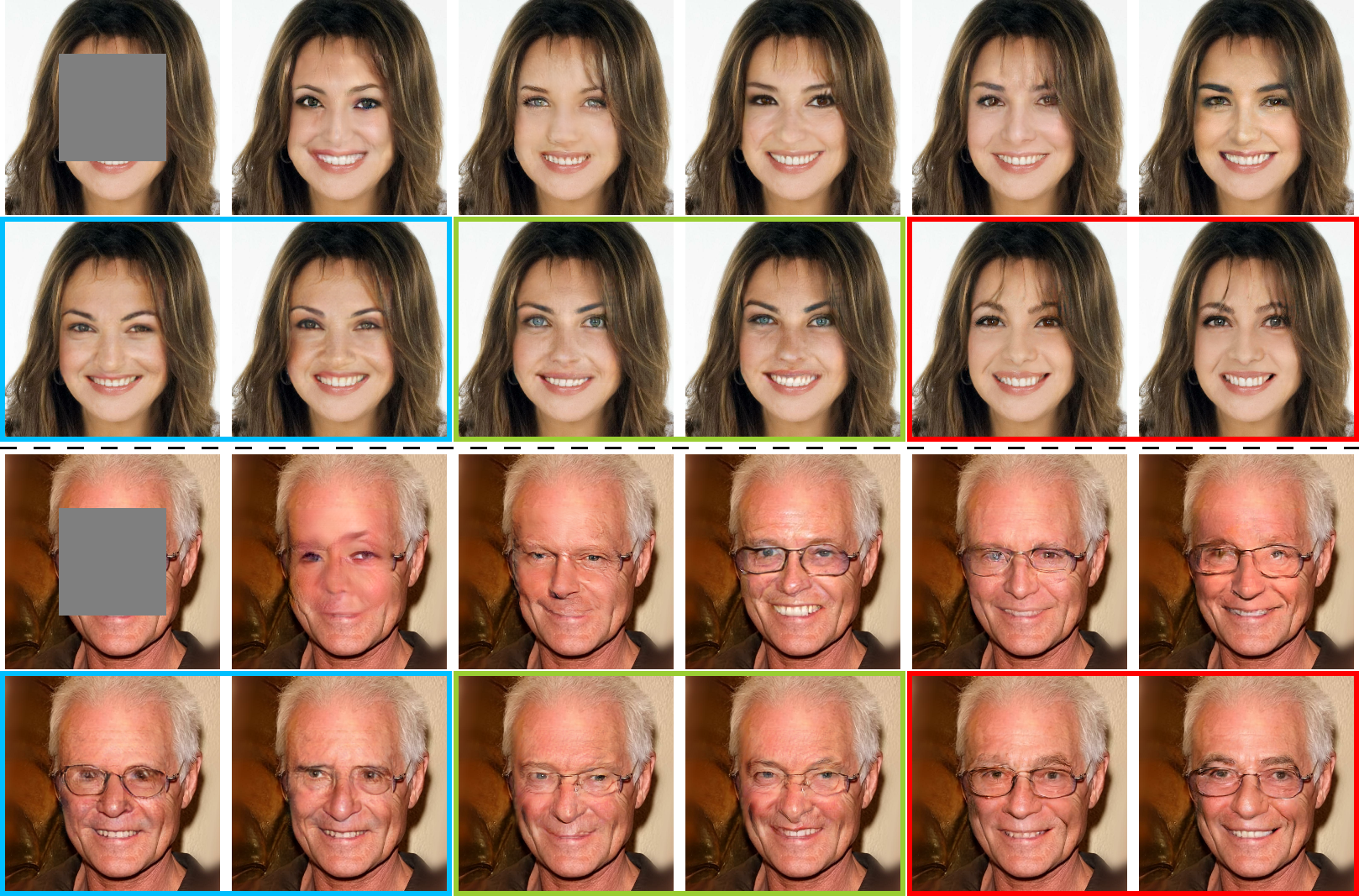}
	\end{center}
	\vspace{-0.24cm}
	\caption{Qualitative comparison results on two test images of CelebA-HQ. For each group, from top to bottom, from left to right, the pictures are: incomplete image, results of CA \cite{yu2018generative}, GC \cite{yu2019free}, CSA \cite{liu2019coherent}, SF \cite{ren2019structureflow}, FE \cite{liu2020rethink}, results of PIC \cite{zheng2019pluralistic} (with blue box), results of UCTGAN \cite{zhao2020uctgan} (with green box), and results of our method (with red box).}
	\label{fig3}
	\vspace{+0.2cm}
\end{figure*}
\begin{figure*}
	\begin{center}
		\includegraphics[width=0.83\linewidth]{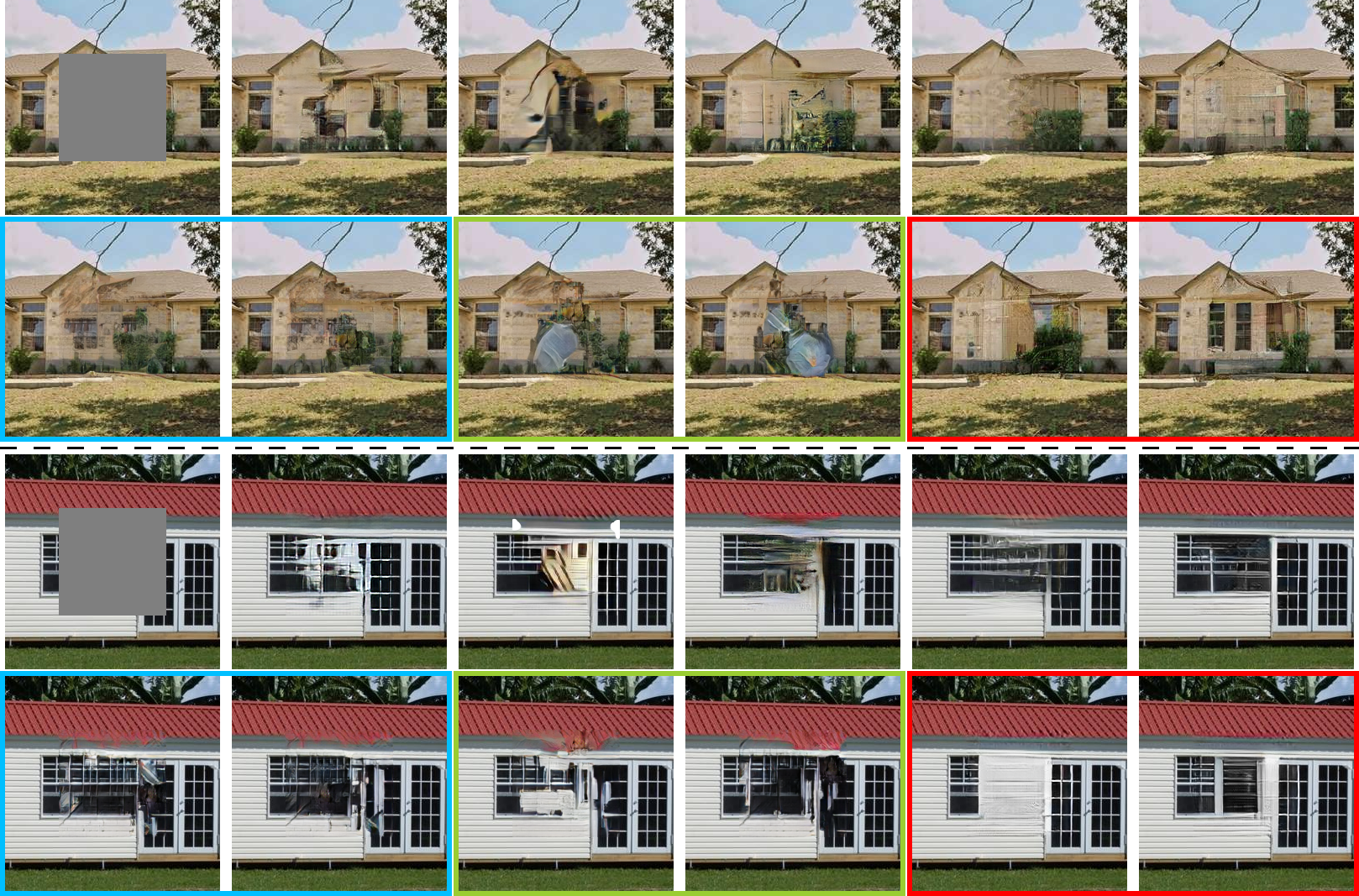}
	\end{center}
	\vspace{-0.24cm}
	\caption{Qualitative comparison results on two test images of Places2. For each group, from top to bottom, from left to right, the pictures are: incomplete image, results of CA \cite{yu2018generative}, GC \cite{yu2019free}, CSA \cite{liu2019coherent}, SF \cite{ren2019structureflow}, FE \cite{liu2020rethink}, results of PIC \cite{zheng2019pluralistic} (with blue box), results of UCTGAN \cite{zhao2020uctgan} (with green box), and results of our method (with red box).}
	\label{fig4}
	\vspace{-0.4cm}
\end{figure*}
\begin{table*}[t]
	\begin{center}
		\begin{tabular}{clccccc}
			\toprule
			\multicolumn{2}{c}{Method} & PSNR$^\uparrow$& SSIM$^\uparrow$ & IS$^\uparrow$ & MIS$^\uparrow$ & FID$^\downarrow$ \\
			\midrule
			\multirow{5}{*}{\tabincell{l}{Single-\\ Solution}} &CA \cite{yu2018generative} & 23.65  & 0.8525 & 3.206 & 0.0207 & 16.64 \\
			&GC \cite{yu2019free} & 25.23  & 0.8713 & 3.384 & 0.0235 & 12.24  \\
			&CSA \cite{liu2019coherent} & \textbf{25.26}  & \textbf{0.8840} & \textcolor{red}{3.408} & 0.0199 & 11.78 \\
		    &SF \cite{ren2019structureflow} & 25.05  & 0.8717 &  3.360 & 0.0229 & \textcolor{red}{10.59} \\
			&FE \cite{liu2020rethink} & 24.10  & 0.8632 & 3.357 & \textcolor{red}{0.0240}  & 10.73 \\
			\midrule
			\multirow{3}{*}{\tabincell{l}{Multiple-\\ Solution}}&PIC \cite{zheng2019pluralistic} & 23.93  & 0.8567 & 3.357 & 0.0225 & 11.70 \\
			&UCTGAN \cite{zhao2020uctgan} &  24.39 & 0.8603 & 3.342 & 0.0237 & 11.74 \\
			&Ours & \textcolor{red}{24.56} & \textcolor{red}{0.8675} & \textbf{3.456} & \textbf{0.0245} & \textbf{9.784} \\
			\bottomrule
		\end{tabular}
	\end{center}
	\vspace{-0.1cm}
	\caption{Quantitative comparison of different methods on the CelebA-HQ test set. For multiple-solution methods, we sample 50 images for each incomplete image and report the average result. Note that PSNR and SSIM are full-reference metrics that compare the generated image with the ground truth, but IS, MIS, and FID are not. For each metric, the best score is highlighted in \textbf{bold}, and the best score within the other category is highlighted in \textcolor{red}{red}.}
	\label{table1}
	\vspace{-0.25cm}
\end{table*}
Some inpainting methods such as \cite{liu2018image, liu2020rethink} use the pre-trained VGG-16 as an auxiliary evaluator to improve the perceptual quality of the results. They define a perceptual loss and a style loss based on the VGG features to train the generator. Inspired by these feature losses, we propose two novel feature losses by reusing our pre-trained hierarchical encoder $E_{vq}$ as an auxiliary evaluator. As shown in Figure \ref{fig2}, $E_{vq}$ maps ${\rm \mathbf{I}_{comp}}$ onto structural features $\textbf{s}_{\rm comp}$ and textural features $\textbf{t}_{\rm comp}$. The structural feature loss of $G_t$ is defined as the multi-class cross-entropy between $\textbf{s}_{\rm comp}$ and $\bar{\textbf{s}}_{\rm gt}$:
\begin{equation}
	\mathcal{L}_{sf}=-\sum_{i,j}I_{ij}\,{\rm log}({\rm softmax}(\lambda_2\tilde{d}_{ij}))
	\vspace{-0.1cm}
\end{equation}
Here, we set $\lambda_2$ = 10. $\tilde{d}_{ij}$ denotes the truncated distance similarity score between the $i^{th}$ feature vector of $\textbf{s}_{\rm comp}$ and the $j^{th}$ prototype vector of the structural codebook. $I_{ij}$ is an indicator of the prototype vector class. $I_{ij}$ = 1 when the $i^{th}$ feature vector of $\bar{\textbf{s}}_{\rm gt}$ belongs to the $j^{th}$ class of the structural codebook, otherwise $I_{ij}$ = 0. The textural feature loss (denoted as $\mathcal{L}_{tf}$) is similar to $\mathcal{L}_{sf}$. The total loss of $G_t$ is defined as:
\begin{equation}
	\mathcal{L}_{tg}=\alpha_{\ell1}\mathcal{L}_{\ell1}+\alpha_{adv}\mathcal{L}_{adv}+\alpha_{f}(\mathcal{L}_{sf}+\mathcal{L}_{tf})
\end{equation}
where $\alpha_{\ell1}$, $\alpha_{adv}$, and $\alpha_f$ are loss weights.
\section{Experiments}
\subsection{Implementation Details}
Our model is implemented in TensorFlow v1.12 and trained on two NVIDIA 2080 Ti GPUs. The batch size is 8. During optimization, the weights of different losses are set to $\alpha_{\ell_1}$ = $\alpha_{\ell_2}$ = $\alpha_{adv}$ = 1, $\alpha_c$ = 0.25, $\alpha_{f}$ = 0.1. We use the Adam optimizer to train the three parts of our model. The learning rate of $E_{vq}$-$D_{vq}$ is $10^{-4}$. The learning rate of $G_s$ follows the linear
warm-up and square-root decay schedule used in \cite{razavi2019generating}. The learning rate of $G_t$ is $10^{-4}$ and $\beta_1$ = 0.5.  We also use polyak exponential moving average (EMA) decay of 0.9997 when training $E_{vq}$-$D_{vq}$ and $G_s$. Each part is trained for 1M iterations. During training, $E_{vq}$-$D_{vq}$ is firstly trained, and then $G_s$ and $G_t$ are trained individually. During inference, only $G_s$ and $G_t$ are used.
\subsection{Performance Evaluation}
We evaluate our method on three datasets including CelebA-HQ \cite{karras2017progressive}, Places2 \cite{zhou2017places}, and ImageNet \cite{russakovsky2015imagenet}. We use the original training, testing, and validation splits for these three datasets. For CelebA-HQ, training images are down-sampled to 256$\times$256 and data augmentation is adopted. For Places2 and ImageNet, training images are randomly cropped to 256$\times$256. The missing regions of the incomplete images can be regular or irregular. We compare our method with state-of-the-art single-solution and multiple-solution inpainting methods. The single-solution methods among them are CA \cite{yu2018generative}, GC \cite{yu2019free}, CSA \cite{liu2019coherent}, SF \cite{ren2019structureflow} and FE \cite{liu2020rethink}. The multiple-solution methods among them are PIC \cite{zheng2019pluralistic} and UCTGAN \cite{zhao2020uctgan}

\textbf{Qualitative Comparisons.} Figure \ref{fig3} and Figure \ref{fig4} show the qualitative comparison results of center-hole inpainting on CelebA-HQ and Places2, respectively. It is difficult for CA \cite{yu2018generative}, GC \cite{yu2019free} and CSA \cite{liu2019coherent} to generate reasonable structures without structural information acts as prior knowledge. SF \cite{ren2019structureflow} and FE \cite{liu2020rethink} use edge-preserved smooth images to guide structure generation. However, they struggle to synthesize fine-grained textures, which indicates that their structural information provides limited help to texture generation. PIC \cite{zheng2019pluralistic} and UCTGAN \cite{zhao2020uctgan} show high diversity. But their results are of low quality, especially for the challenging Places2 test images. Compared to these methods, the results of our method have more reasonable structures and more realistic textures, \eg fine-grained hair and eyebrows in Figure \ref{fig3}. In addition, the diversity of our method is enhanced, \eg different eye colors in Figure \ref{fig3} and varying window sizes in Figure \ref{fig4}. More qualitative results and analyses of artifacts are presented in the supplementary material.



\textbf{Quantitative Comparisons.} Following previous image inpainting methods, we use common evaluation metrics such as peak signal-to-noise ratio (PSNR) and structural similarity (SSIM) to measure the similarity between the inpainting result and ground truth. However, these full-reference metrics are not suitable for the image inpainting task because there are multiple plausible solutions for an incomplete image. The inpainting methods are supposed to focus on generating realistic results rather than merely approximating ground truth. Therefore, we also use Inception Score (IS) \cite{salimans2016improved}, Modified Inception Score (MIS) \cite{zhao2020uctgan}, and Fréchet Inception Distance (FID) \cite{heusel2017gans} as perceptual quality metrics. These metrics are consistent with human judgment \cite{heusel2017gans}. FID can also detect GAN’s known shortcomings such as mode collapse and mode dropping \cite{lucic2018gans}.

Unlike previous multiple-solution methods \cite{zheng2019pluralistic, zhao2020uctgan} that use discriminator to select samples for quantitative evaluation, we use all samples for fair comparison. The comparison is conducted on CelebA-HQ 1000 testing images with 128$\times$128 center holes. As shown in Table \ref{table1}, multiple-solution methods score relatively low on PSNR and SSIM because they generate highly diverse results instead of approximating ground truth. Still, our method outperforms PIC \cite{zheng2019pluralistic} and UCTGAN \cite{zhao2020uctgan} on these two metrics. Furthermore, our method outperforms all the other methods in terms of IS, MIS and FID.

We also evaluate the diversity of our method using the LPIPS \cite{zhang2018unreasonable} metric. The average score is calculated between consecutive pairs of 50K results which are sampled from 1K incomplete images. Higher score indicates higher diversity. The reported scores of PIC \cite{zheng2019pluralistic} and UCTGAN \cite{zhao2020uctgan} are 0.029 and 0.030, respectively. Our method achieves a comparable score of 0.029. Please refer to the supplementary material for some discussions about the diversity.

\subsection{Ablation Study}\label{sec4.3}
\begin{figure}[t]
	\begin{center}
		\includegraphics[width=0.825\linewidth]{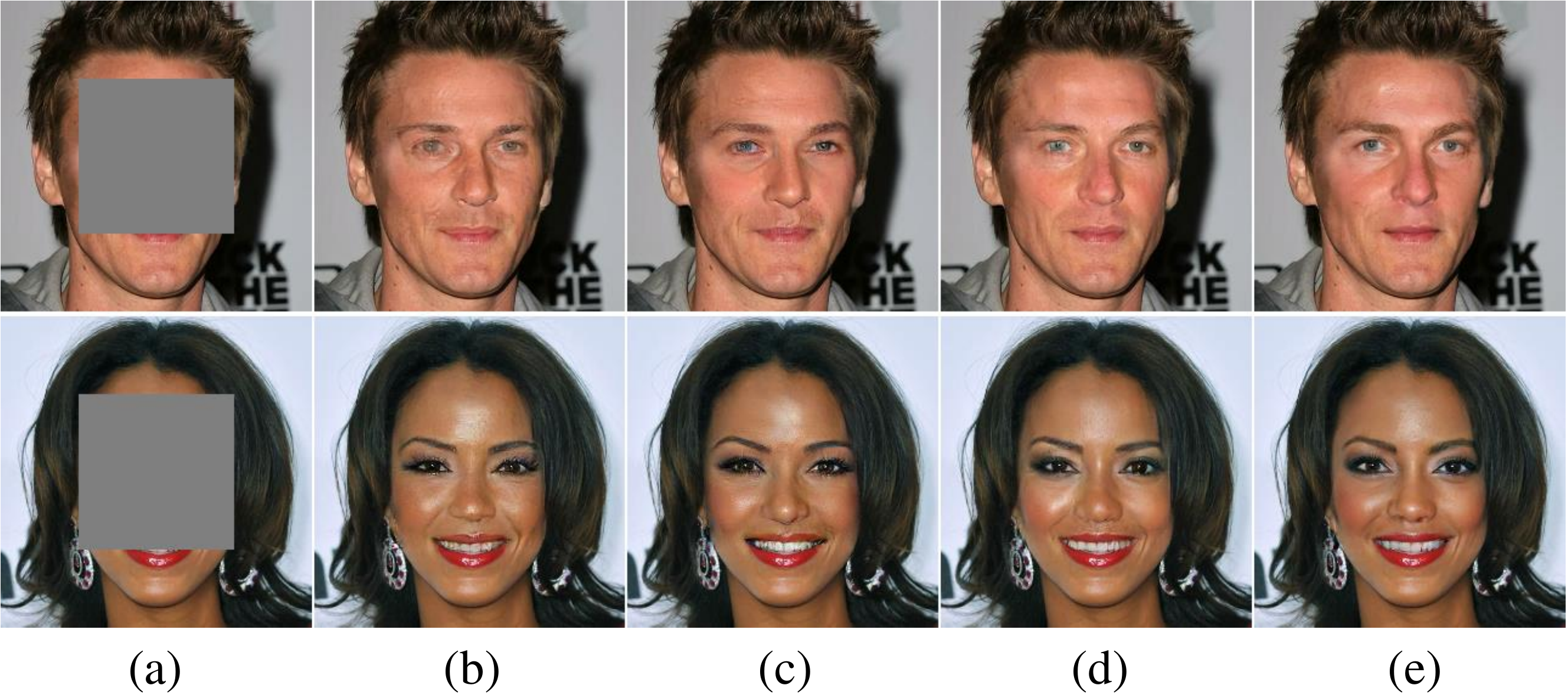}
	\end{center}
	\vspace{-0.29cm}
	\caption{Results of the ablation study on the structural attention module. (a) Incomplete image. (b) Using cross attention on the learned features. (c) Using full attention on the learned features. (d) Using cross attention on the structural features. (e) Using full attention on the structural features. [Best viewed with zoom-in.]}
	\label{fig6}
	\vspace{-0.10cm}
\end{figure}
\begin{table}
	\begin{center}
		\begin{tabular}{p{1.9cm}<{\centering}p{1.8cm}ccc}
			\toprule
			Full & Structural &  SSIM$^\uparrow$ & IS$^\uparrow$ & FID$^\downarrow$ \\
			\midrule
			&    & 0.8606 & 3.402 & 11.55 \\
			\checkmark&    &0.8645 & 3.436& 11.26\\
			& $\quad\ \ $\checkmark   &0.8675 & 3.416& 10.12\\
			\checkmark& $\quad\ \ $\checkmark   & \textbf{0.8676} & \textbf{3.467} & \textbf{9.670}\\
			\bottomrule
		\end{tabular}
	\end{center}
	\vspace{-0.1cm}
	\caption{Quantitative comparison for the ablation study on the structural attention module. Refer to Section \ref{sec4.3} and Figure \ref{fig6} for description. }
	\label{table2}
	\vspace{-0.25cm}
\end{table}
We conduct an ablation study on CelebA-HQ to show the effect of different components of the texture generator $G_t$. Since the structure generator $G_s$ can produce diverse structural features, we randomly sample a set of generated structural features. Then we use it across all the following experiments for fair comparison.

\textbf{Effect of structural attention module.} We compare the effect of different attention modules. The attention module in \cite{yi2020contextual} calculates cross attention scores on the learned features, which often results in texture artifacts (see Figure \ref{fig6}(b)). Using full attention instead of cross attention maintains global consistency (see Figure \ref{fig6}(c)). Using the structural features instead of the learned features improves the consistency between structures and textures (see Figure \ref{fig6}(d)). Our structural attention module calculates full attention scores on the structural features, which can synthesize realistic textures, such as symmetric eyes and eyebrows (see Figure \ref{fig6}(e)). The quantitative results in Table \ref{table2} also demonstrate the benefits of our structural attention module.



\begin{figure}[t]
	\begin{center}
		\includegraphics[width=1.0\linewidth]{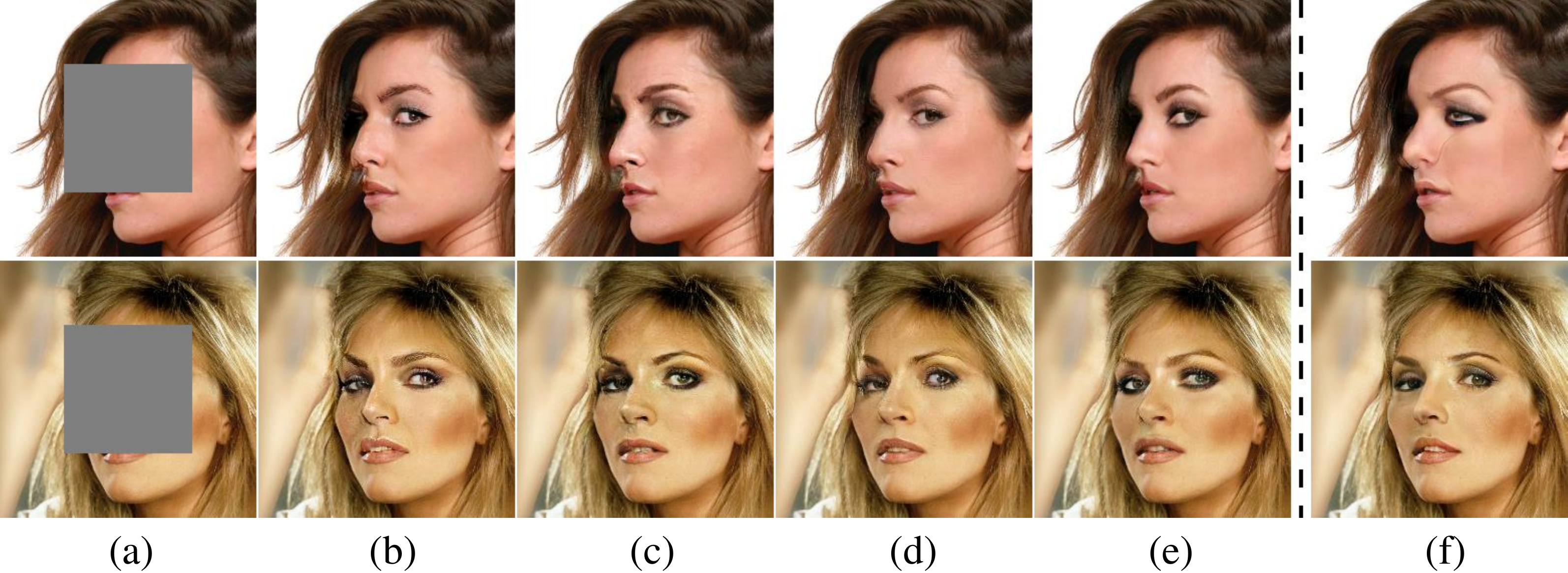}
	\end{center}
	\vspace{-0.28cm}
	\caption{Results of the ablation study on the auxiliary losses. (a) Incomplete image. (b) Using no feature losses (but still using $\mathcal{L}_{\ell1}$ and $\mathcal{L}_{adv}$). (c) Using $\mathcal{L}_{sf}$. (d) Using $\mathcal{L}_{tf}$. (e) Using both $\mathcal{L}_{sf}$ and $\mathcal{L}_{tf}$. (f) Using the perceptual and style losses as in \cite{liu2018image, liu2020rethink}. [Best viewed with zoom-in.]}
	\label{fig7}
	\vspace{-0.05cm}
\end{figure}

\textbf{Effect of our feature losses.} We compare the effect of different auxiliary losses. Without any auxiliary loss, the perceptual quality of the inpainting result is not satisfactory (see Figure \ref{fig7}(b)). Using our structural feature loss $\mathcal{L}_{sf}$ can improve structure coherence, such as the shape of nose and mouth (see Figure \ref{fig7}(c)). Using our textural feature loss $\mathcal{L}_{tf}$ can improve texture realism, such as the luster of facial skin (see Figure \ref{fig7}(d)). Using both $\mathcal{L}_{sf}$ and $\mathcal{L}_{tf}$ can generate more natural images (see Figure \ref{fig7}(e)). Compared to our feature losses, perceptual and style losses used in \cite{liu2018image, liu2020rethink} may produce a distorted structure or an inconsistent texture (see Figure \ref{fig7}(f)). The quantitative results in Table \ref{table3} also demonstrate the benefits of our feature losses.\vspace{-0.10cm}

\begin{table}
	\begin{center}
		\begin{tabular}{p{1.9cm}<{\centering}p{1.2cm}<{\centering}ccc}
			\toprule
			$\mathcal{L}_{sf}$&$\mathcal{L}_{tf}$ & SSIM$^\uparrow$ & IS$^\uparrow$ & FID$^\downarrow$ \\
			\midrule
			&  & 0.8581 & 3.383 & 11.02\\
			\checkmark & & 0.8589 & 3.367& 11.16\\
			&\checkmark & 0.8625 & 3.414& 10.34\\
			\checkmark& \checkmark & \textbf{0.8676} & \textbf{3.467} & \textbf{9.670}\\
			\midrule
		    \multicolumn{2}{c}{$\mathcal{L}_{perceptual}$ + $\mathcal{L}_{style}$ \cite{liu2018image, liu2020rethink}} & 0.8638 &3.388 & 9.672\\
			\bottomrule
		\end{tabular}
	\end{center}
	\vspace{-0.1cm}
	\caption{Quantitative comparison for the ablation study on the auxiliary losses. Refer to Section \ref{sec4.3} and Figure \ref{fig7} for description.}
	\label{table3}
	\vspace{-0.45cm}
\end{table}
\section{Conclusion}
We have proposed a multiple-solution inpainting method for generating diverse and high-quality images using hierarchical VQ-VAE. Our method first formulates an autoregressive distribution to generate diverse structures, then synthesizes the image texture for each kind of structure. We propose a structural attention module to ensure that the synthesized texture is consistent with the generated structure. We further propose two feature losses to improve structure coherence and texture realism, respectively. Extensive qualitative and quantitative comparisons show the superiority of our method in both quality and diversity. We demonstrate that the structural information extracted by the hierarchical VQ-VAE is of great benefit for the inpainting task.  
As for future work, we plan to extend our method to other conditional image generation tasks including style transfer, image super-resolution, and guided editing.

\clearpage
{\small
\bibliographystyle{ieee_fullname}
\bibliography{egbib}
}

\renewcommand\thesection{\Alph{section}}
\setcounter{section}{0}
\section{Architecture Hyperparameters and Training Details.}
The hyperparameters used for training the hierarchical VQ-VAE are reported in Table \ref{supp_table1}. The hyperparameters used for training the diverse structure generator are reported in Table \ref{supp_table2}.  As for the GAN-based texture generator, the hidden units of generator and discriminator are both 64. Our model is implemented in TensorFlow v1.12. Batch size is 8. We train the hierarchical VQ-VAE and the texture generator on a single NVIDIA 2080 Ti GPU, and train the diverse structure generator on two GPUs. Each part is trained for 10$^6$ iterations. Training the hierarchical VQ-VAE takes roughly 8 hours. Training the diverse structure generator takes roughly 5 days. Training the texture generator takes roughly 4 days. All the training time is independent of the data set and type of masks. Note that the diverse structure generator and the texture generator can be trained in parallel after the training of the hierarchical VQ-VAE.

\section{Negative Log Likelihood and Reconstruction Error.} The VQ-VAE is inspired by lossy compression where performance is usually characterized with rate-distortion curves \cite{razavi2019generating}. Our hierarchical VQ-VAE minimizes the mean-square-error (MSE) reconstruction error as the distortion metric, while our diverse structure generator minimizes the negative log likelihood (NLL) of global latent. As such, we report the distortion in MSE and the NLL of global latent (estimate of coding rate) in Table \ref{supp_table3}. We do not measure the NLL of local latent because we use a GAN rather than likelihood-based network for texture generation. Note that NLL values are only comparable
between likelihood-based networks that use the same pre-trained VQ-VAE. Our diverse structure generator retains the advantages of likelihood-based methods, such as a clear objective to compare models, progress tracking, and measurement of over-fitting and mode coverage (the properties that result in diverse samples).

\section{Inference Time.} One advantage of GAN-based and VAE-based methods is their fast inference speed. We measure that FE \cite{liu2020rethink} runs at 0.2 second per image on a single NVIDIA 1080 Ti GPU for images of resolution 256$\times$256. In contrast, our model runs at 45 seconds per image. Naively sampling our autoregressive network is the major source of computational time. Fortunately, this time can be reduced by an order of magnitude using an incremental sampling technique \cite{ramachandran2017fast} which caches and reuses intermediate states of the network. We may integrate this technique in the future.

\section{More Visual Examples.} Following the recent inpainting methods, we use center masks or random masks to train our models on the CelebA-HQ, Places2, and ImageNet datasets. The center masks are 128$\times$128 center holes in the 256$\times$256 images. The random masks are rectangles and brush strokes with random positions and sizes, similar to those in \cite{zheng2019pluralistic}. We first show more results of our method using the center-mask models (see Figures \ref{fig_a21}, \ref{fig_a22} and \ref{fig_a23}) and the random-mask models (see Figure \ref{fig_a24}). We also show some failure cases of our method (see Figure \ref{fig_a25}). Then we show that the degree of diversity is controlled by the condition, \ie the available content (see Figures \ref{fig_a26} and \ref{fig_a27}). The degree of diversity is also  controlled by the location and size of the missing region (see Figure \ref{fig_b1}). 


\section{Discussions on Artifacts.} 
Although our method can generate more realistic results than prior works, some results still have noticeable artifacts. We analyze the reasons of these artifacts and find that most of them are due to the low quality of the generated structures. Note that we used a light-weight autoregressive network in the structure generator for the sake of computational efficiency, compared with the original, much more complex network in \cite{razavi2019generating}. We anticipate that the results will be improved if using the complex network. In addition, our texture generator also incurs some artifacts. We may improve it by integrating the new techniques proposed in the recent single-solution inpainting studies, such as feature discriminator \cite{liu2019coherent}, multi-scale discriminator \cite{ren2019structureflow}, and multi-scale generator \cite{liu2020rethink}.

\section{Discussions on Diversity.} 
\begin{figure*}
	\begin{center}
		\includegraphics[width=0.98\linewidth]{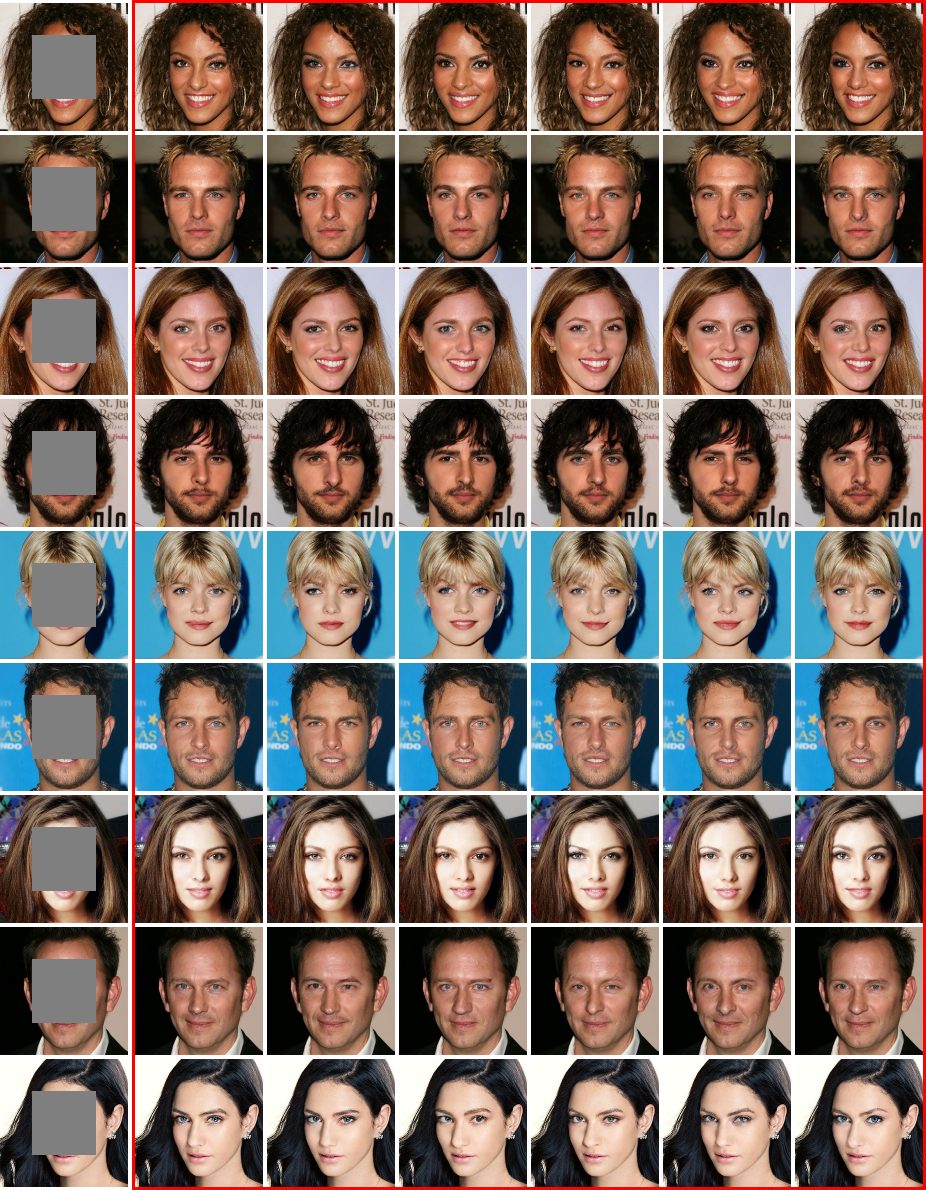}
	\end{center}
	\vspace{-0.24cm}
	\caption{Additional results on the CelebA-HQ test set using the center-mask CelebA-HQ model.}
	\label{fig_a21}
\end{figure*}

\begin{figure*}
	\begin{center}
		\includegraphics[width=0.98\linewidth]{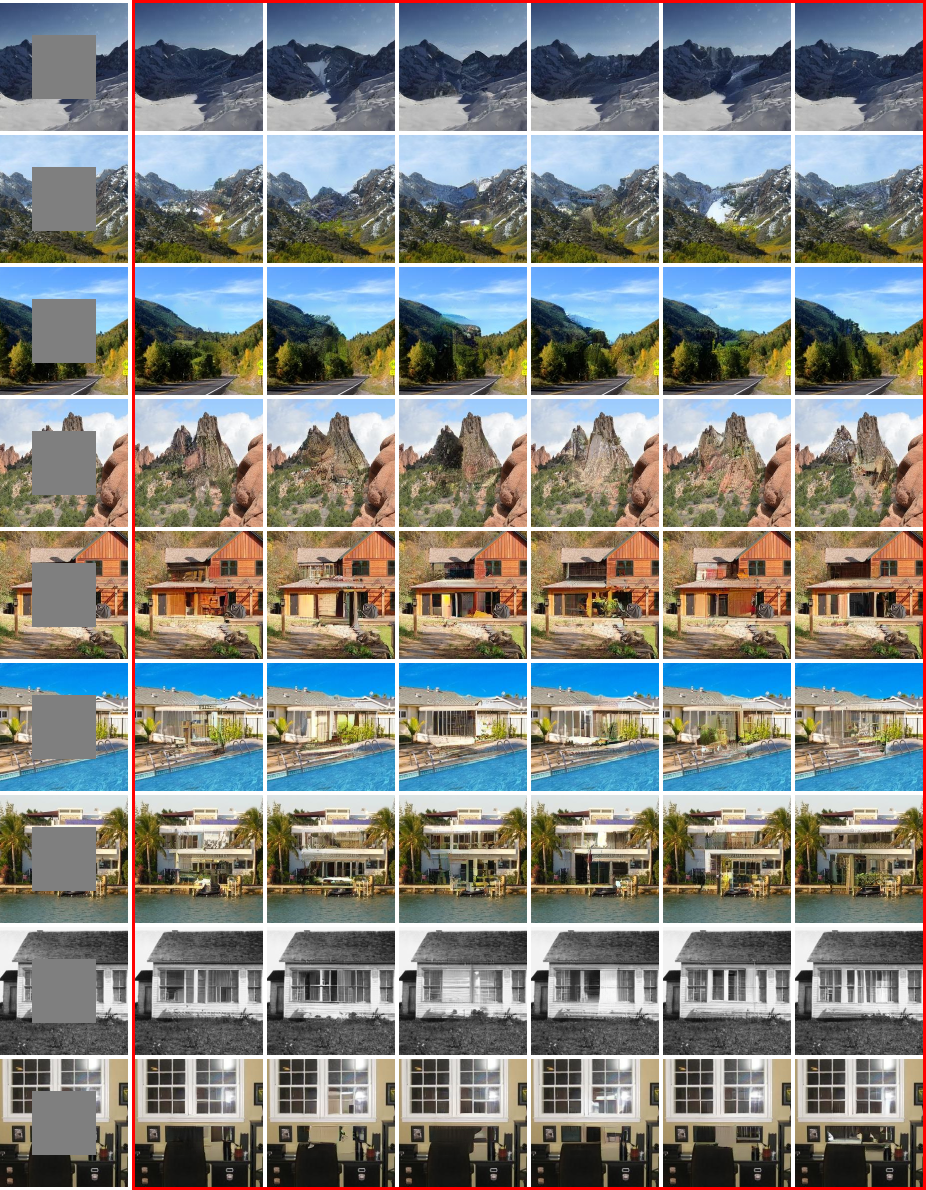}
	\end{center}
	\vspace{-0.24cm}
	\caption{Additional results on the Places2 validation set using the center-mask Places2 model. }
	\label{fig_a22}
\end{figure*}

\begin{figure*}
	\begin{center}
		\includegraphics[width=0.98\linewidth]{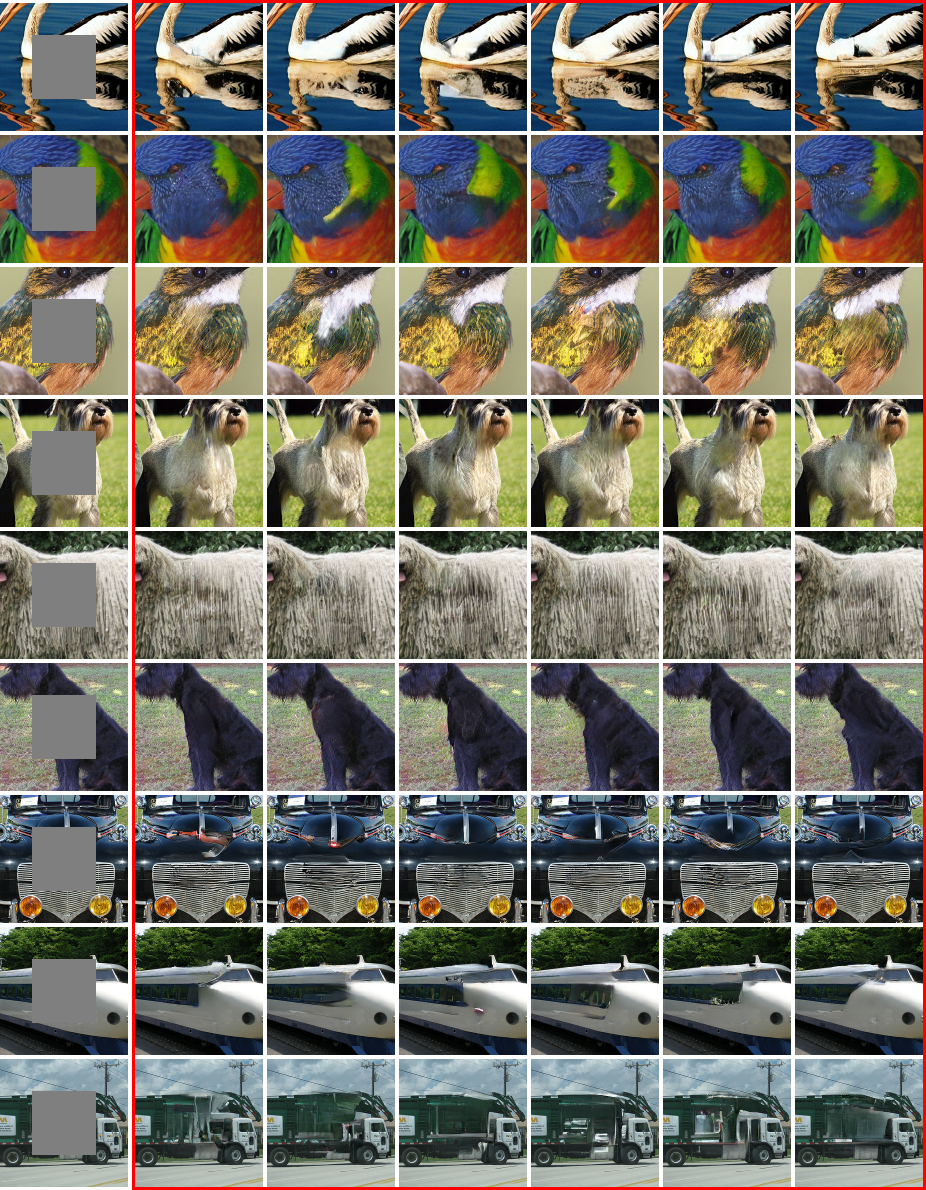}
	\end{center}
	\vspace{-0.24cm}
	\caption{Additional results on the ImageNet validation set using the center-mask ImageNet model.}
	\label{fig_a23}
\end{figure*}

\begin{figure*}
	\begin{center}
		\includegraphics[width=0.98\linewidth]{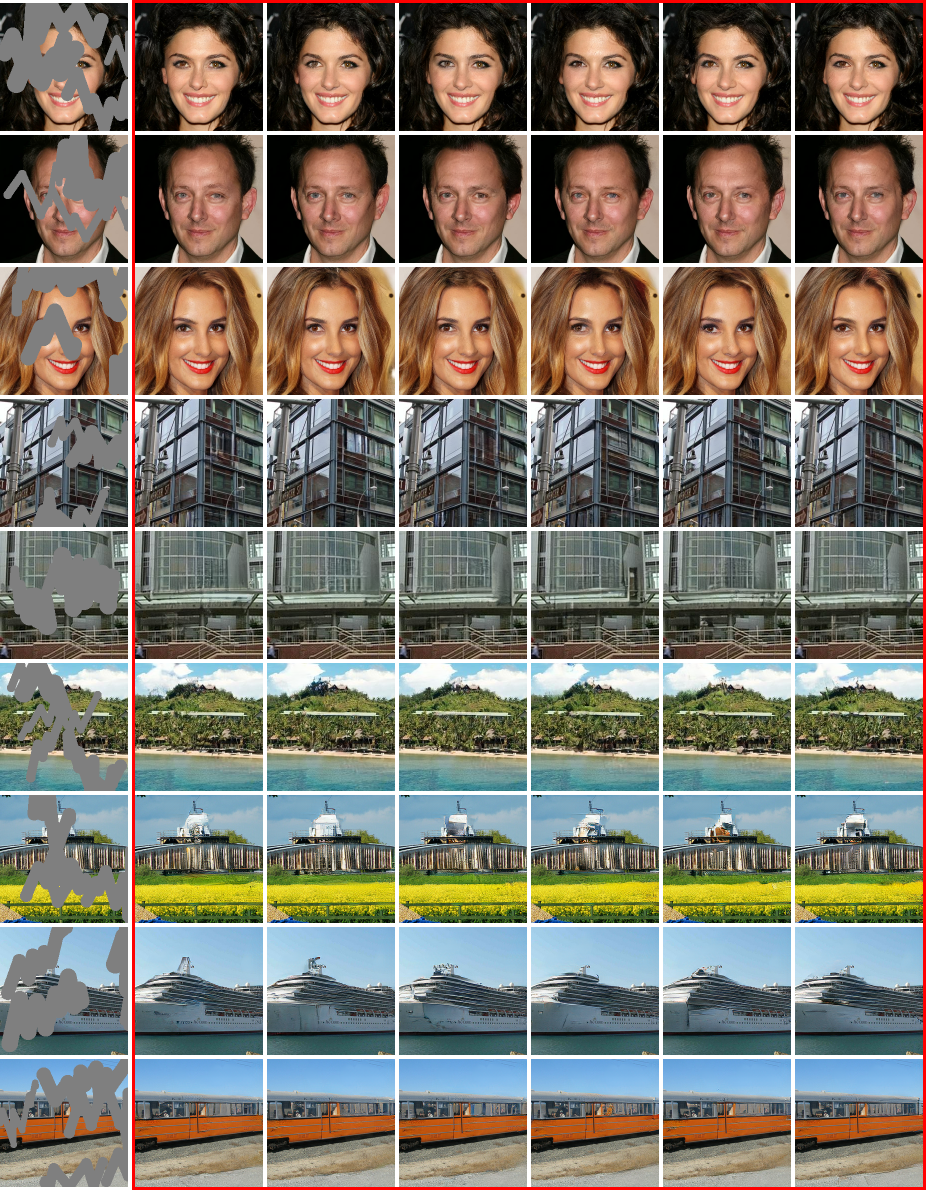}
	\end{center}
	\vspace{-0.24cm}
	\caption{Additional results on the CelebA-HQ, Places2, and ImageNet test (or validation) sets using the random-mask models. }
	
	\label{fig_a24}
\end{figure*}

\begin{figure*}
	\begin{center}
		\includegraphics[width=0.98\linewidth]{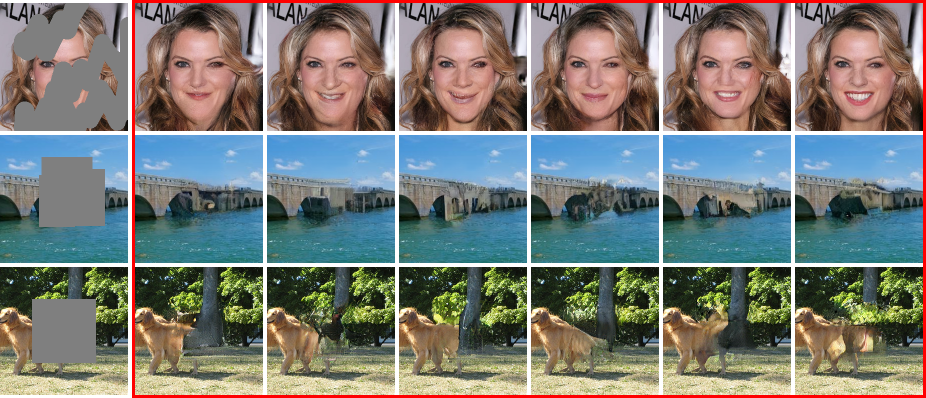}
	\end{center}
	\vspace{-0.20cm}
	\caption{Failure cases of our method on the CelebA-HQ, Places2, and ImageNet test (or validation) sets. (Top) Face image inpainting with big holes. The results are of low quality, \eg distorted faces, asymmetric eyes, missing nostrils, and blurry teeth. (Middle) Scene image inpainting with a complex structure. The results cannot reconstruct the bridge architecture and the bridge holes of varying sizes. (Bottom) Natural image inpainting with a hole of mixed foreground (\ie dog) and background (\ie trees). A large portion of foreground is lost. And the generated results have unsatisfactory structures and blurry textures. } 
	\label{fig_a25}
	\vspace{+0.2cm}
\end{figure*}

\begin{figure*}
	\begin{center}
		\includegraphics[width=0.98\linewidth]{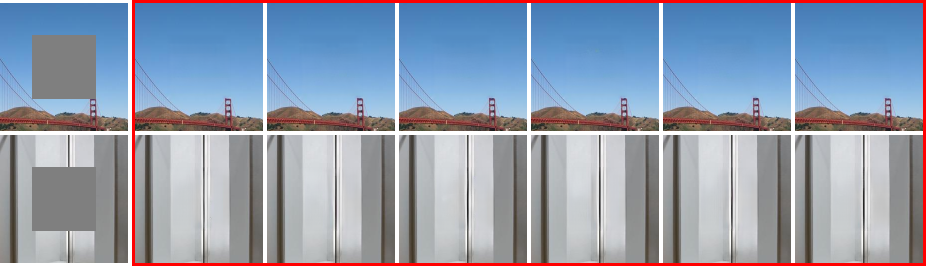}
	\end{center}
	\vspace{-0.24cm}
	\caption{Additional results of our method with low diversity. The degree of diversity is limited when the available content has a simple structure and a plain texture.}
	\label{fig_a26}
	\vspace{+0.2cm}
\end{figure*}

\begin{figure*}
	\begin{center}
		\includegraphics[width=0.98\linewidth]{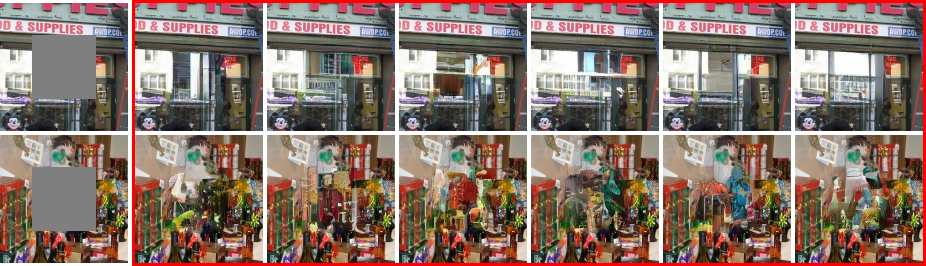}
	\end{center}
	\vspace{-0.24cm}
	\caption{Additional results of our method with high diversity. The degree of diversity is high when the available content has a complex structure and an intricate texture.}
	\label{fig_a27}
\end{figure*}

\begin{figure*}
	\begin{center}
		\includegraphics[width=0.98\linewidth]{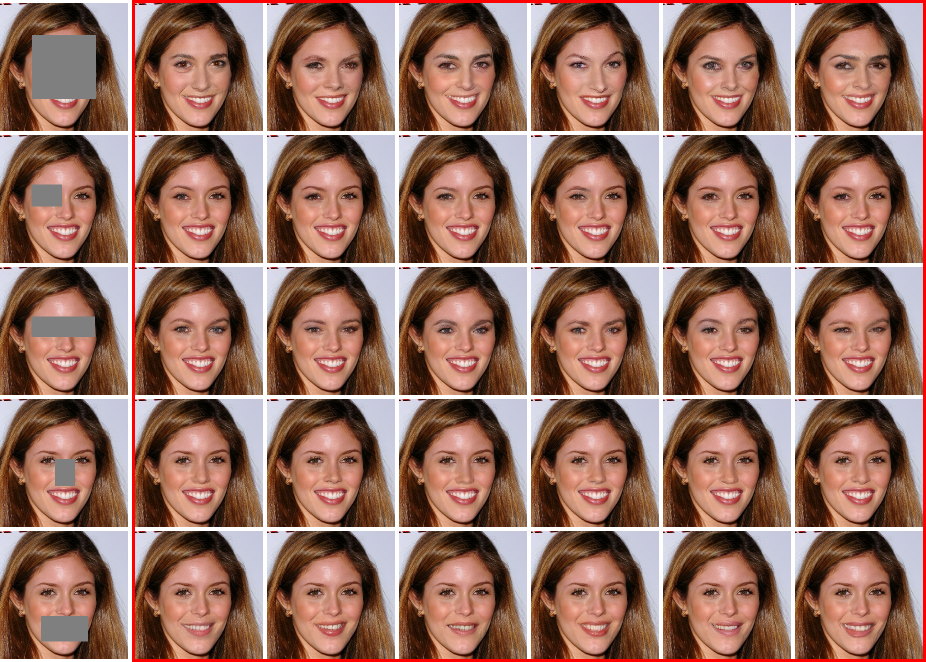}
	\end{center}
	\vspace{-0.20cm}
	\caption{Additional results on one CelebA-HQ test image with different holes using the random-mask CelebA-HQ model. For different rows, the degree of diversity is controlled by the location and size of the missing region.}
	\label{fig_b1}
	\vspace{-0.3cm}
\end{figure*}

\begin{figure*}
	\begin{center}
		\includegraphics[width=0.47\linewidth]{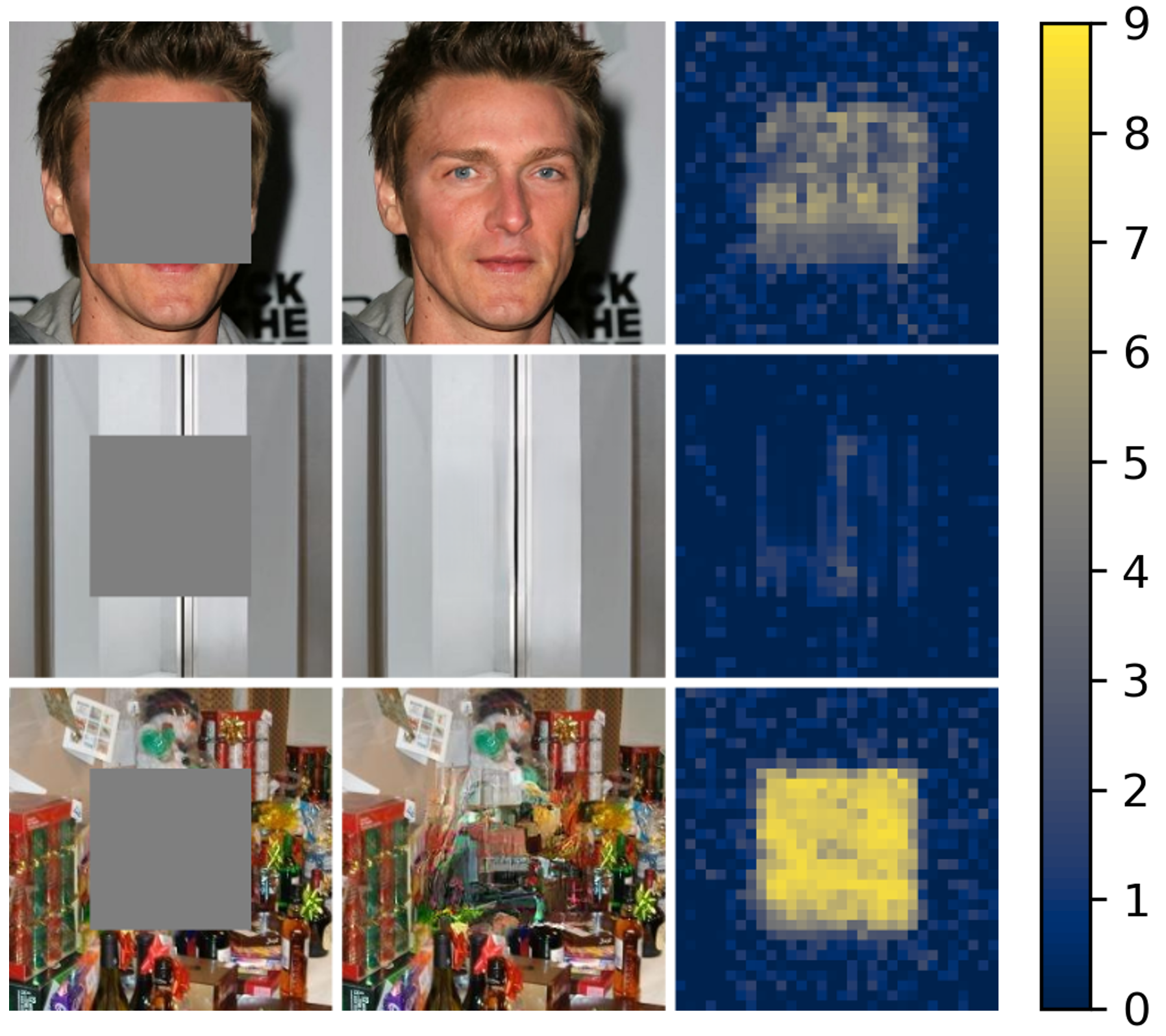}
	\end{center}
	\vspace{-0.38cm}
	\caption{Visualization results of the entropy of learned distribution. For each row, from left to right, the pictures are: incomplete image, one result of our method, and the corresponding visualized entropy. The maximum entropy is 9 because the codebook size is $K$ = 2$^9$. }
	\label{fig_b2}
\end{figure*}

\indent In our method, the diversity is fully determined by the learned conditional distribution for structure generation (since the texture generation has no randomness). We visualize the pixel-wise entropy of the learned distribution to analyze the diversity. As shown in Figure \ref{fig_b2}, the training dataset and the complexity of incomplete image have impact on the entropy. Intuitively, higher entropy leads to higher diversity. \\
\indent The diversity of the inpainting results depends on at least the following factors.\\
\indent (1) \textit{The training dataset.} Since our method learns a conditional distribution for diverse structure generation, it always benefits from diverse training data to enrich the learned distribution. This is evidenced by the experimental results that the resulting diversity on the face dataset is clearly less than that on the natural image datasets (Figure \ref{fig_a21} vs.\ Figure \ref{fig_a22}); note that the face training images ($\sim$10$^4$) are far less than the natural training images ($\sim$10$^7$). We conjecture that using a larger dataset or performing training data augmentation may be helpful to increase diversity.\\
\indent (2) \textit{The incomplete image for inference.} As inpainting is a conditional generation task, the incomplete image acts as the condition or the constraint. The available content in the incomplete image, and the location/size of the missing region, both decide the diversity of the results to a large extent. The effect of the available content is shown in Figure \ref{fig_a26} and Figure \ref{fig_a27}. The effect of the location/size of the missing region is shown in Figure \ref{fig_b1}. \\
\indent (3) \textit{The mask type.} We use center masks or random masks to train our models. We find that the models trained with random masks seem to have higher diversity, even for the same incomplete image (Figure \ref{fig_a21} Row 3 vs.\ Figure \ref{fig_b1} Row 1). We conjecture that using more random masks may be helpful to increase diversity.\\
\indent (4) \textit{The method itself.} Taking our method as example, we may improve the diversity by  increasing the support of the conditional distribution (\eg codebook size), using more sophisticated model for the distribution, adding regularization terms into the loss function (such as to increase the entropy of the distribution), etc.\\
\indent (5) \textit{Diversity-quality tradeoff.} We believe that there may be a tradeoff between diversity and quality in the inpainting task. If we pursue higher diversity, we may try to increase the entropy of the learned distribution
(e.g. by adding regularization terms into the loss function); then, the quality may be deteriorated since the learned distribution is intentionally biased. Moreover, from a broader perspective, inpainting is a signal restoration task; in such tasks there are always different kinds of tradeoff, like the perception-distortion tradeoff \cite{blau2018perception}. We have interest to theoretically study the quality-diversity tradeoff in the future.

\clearpage
\begin{table*}
	\begin{center}
		\begin{tabular}{l|c}
			\toprule
			&$E_{vq}$-$D_{vq}$\\
			\midrule
			Input size &256$\times$256\\
			Latent layers & 32$\times$32, 64$\times$64\\
			Commitment loss weight & 0.25\\
			Batch size &8\\
			Hidden units & 128\\
			Residual units & 64\\
			Layers & 2\\
			Codebook size & 512\\
			Codebook dimension& 64\\
			Conv. filter size &3\\
			Training steps &1,000,000\\
			Polyak EMA decay&0.9997\\
			\bottomrule
		\end{tabular}
	\end{center}
	\vspace{-0.15cm}
	\caption{Hyperparameters of our hierarchical VQ-VAE.}
	\label{supp_table1}
\end{table*}

\begin{table*}
	\begin{center}
		\begin{tabular}{l|c}
			\toprule
			&$G_s$\\
			\midrule
			Input size &256$\times$256\\
			Latent layer & 32$\times$32\\
			Batch size &8\\
			Hidden units & 128\\
			Residual units & 128\\
			Conditioning hidden units& 32\\
			Conditioning residual units& 32\\
			Layers & 20\\
			Attention layers & 4\\
			Attention heads& 8\\
			Conv. Filter size &3\\
			Dropout &0.1\\
			Output stack layers&20\\
			Training steps &1,000,000\\
			Polyak EMA decay&0.9997\\
			\bottomrule
		\end{tabular}
	\end{center}
	\vspace{-0.15cm}
	\caption{Hyperparameters of our diverse structure generator.}
	\label{supp_table2}
\end{table*}

\begin{table*}
	\begin{center}
		\begin{tabular}{l|cccc}
			\toprule
			&Training NLL&Validation NLL&Training MSE&Validation MSE\\
			\midrule
			CelebA-HQ&1.180&1.243&0.0028&0.0033\\
			Places2&0.969&0.952&0.0042&0.0042\\
			ImageNet&1.127&1.113&0.0082&0.0084\\
			\bottomrule
		\end{tabular}
	\end{center}
	\vspace{-0.15cm}
	\caption{Quantitative results of negative log likelihood (NLL) and mean-squared-error (MSE) in the training and validation of our random-mask models. The reported results are evaluated on the CelebA-HQ, Places2, and ImageNet datasets.}
	\label{supp_table3}
\end{table*}

\clearpage

\clearpage

\end{document}